\documentclass[11pt, a4paper]{article} 
\usepackage{fullpage}
\usepackage{crmath}
\usepackage[sort]{cite}
\usepackage{url}
\usepackage[ruled,vlined]{algorithm2e}
\usepackage{graphicx} 
\usepackage{hyperref}
\usepackage{amsthm}
\usepackage{enumitem}

\title{Bayesian Markov Blanket Estimation}

\author{Dinu Kaufmann \qquad Sonali Parbhoo \qquad Aleksander Wieczorek \\ Sebastian Keller \qquad David Adametz \qquad Volker Roth \\ University of Basel, Switzerland \\ \texttt{ \{dinu.kaufmann, sonali.parbhoo, aleksander.wieczorek} \\ \texttt{sebastianmathias.keller, david.adametz, volker.roth\} @ unibas.ch } }

\newtheorem{thm}{Theorem}
\newtheorem{lemma}{Lemma}

\newenvironment{thmenum}
 {\begin{enumerate}[label=\upshape(\arabic*),ref=\thethm(\arabic*)]}
 {\end{enumerate}}

\newenvironment{proofsketch}{%
  \proof}{\endproof}

\begin{document}

\maketitle

\begin{abstract}

This paper considers a Bayesian view for estimating a sub-network in a Markov random field. The sub-network corresponds to the Markov blanket of a set of query variables, where the set of potential neighbours here is big.  We factorize the posterior such that the Markov blanket is conditionally independent of the network of the potential neighbours. By exploiting this blockwise decoupling, we derive analytic expressions for posterior conditionals. Subsequently, we develop an inference scheme which makes use of the factorization. As a result, estimation of a sub-network is possible without inferring an entire network.  Since the resulting Gibbs sampler scales linearly with the number of variables, it can handle relatively large neighbourhoods. The proposed scheme results in faster convergence and superior mixing of the Markov chain than existing Bayesian network estimation techniques.

\end{abstract}

\section{Introduction and Related Work}
Estimating a network of dependencies among a set of objects is a difficult problem in statistics, particularly in high-dimensional settings or where the observed measurements are noisy. Gaussian Graphical Models (GGM) are a tool for representing such relationships in an interpretable way. In a classical GGM setting, the sparsity pattern of the inverse covariance matrix $\ma{W} = \ma{\Sigma}^{-1}$ encodes conditional independence between variables of the graph. Consequently, various estimators have been proposed that reduce the number of parameters by imposing sparsity constraints on $\ma{W}$. Among these, the popular graphical lasso procedure \cite{friedman2008sparse, meinshausen2006high} places a Wishart likelihood on the sample covariance and computes a point estimate of the graph by minimizing the penalised log-likelihood.

In situations where the variables in the network have different types, it is often more interesting to examine the connections between these types as opposed to estimating an entire network of all the associations. Consider the example in gene analysis where the dependency between only a few clinical factors and thousands of genetic markers is required. When we would like to focus on a particular portion of the network, it is useful to limit the estimate to the \textit{Markov blanket} of the nodes we are interested in. These are the set of nodes that, when conditioned on, render the nodes of interest independent of the rest of the network. 

\begin{figure}[!htbp]
  \centering
  \includegraphics[width = \textwidth]{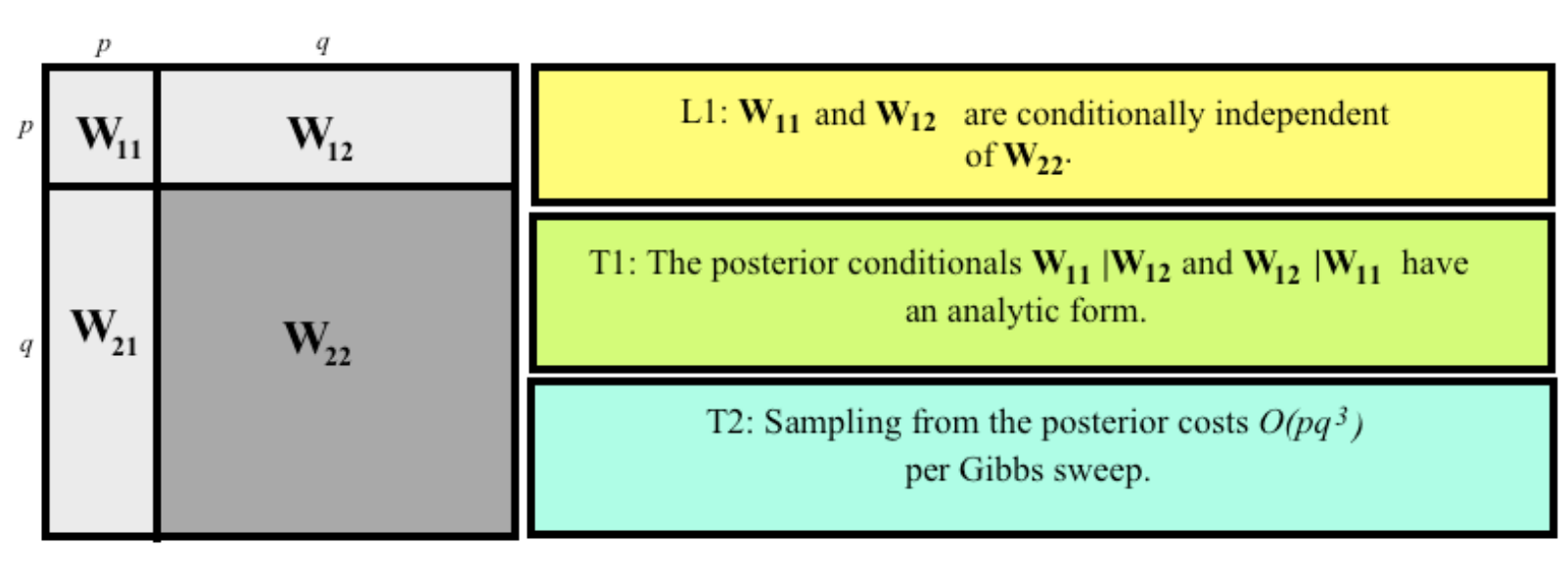}
  \caption{Overview of Bayesian Markov blanket estimation and key results.}
  \label{blockwise}
\end{figure}

In this paper we provide a Bayesian perspective of estimating the Markov blanket of a set of $p$ query variables in an undirected network\footnote{Also known as a \textit{Markov network} or a \textit{Markov random field}.}. Unlike the point estimate of the graphical lasso, the Bayesian view enables the computation of a posterior distribution of the Markov blanket. A Bayesian interpretation of the graphical lasso is presented by Wang \cite{wang2012bayesian}. This approach partitions the matrix $\ma{W}$ as shown on the left in Fig.\,\ref{blockwise}. Posterior inference involves iterating through the individual variables to estimate the entire network. In particular, inference of the $\ma{W}_{12}$ block relies on estimating both $\ma{W}_{11}$ and $\ma{W}_{22}$. However, the coupling of $\ma{W}_{12}$ and $\ma{W}_{22}$ is a limiting factor that can be avoided in the context of Markov blanket estimation. This idea forms the basis of our paper. An important observation for the model we present here, is that the Wishart likelihood may be factorised such that the blocks $\ma{W}_{11}$ and $\ma{W}_{12}$ are de-coupled from $\ma{W}_{22}$. This result is provided as Lemma 1 in Section \ref{sec:cgl}. We show that by combining the factorised likelihood with an appropriate choice of prior, we obtain a posterior distribution that preserves this independence structure. Most importantly, this posterior distribution has an analytic form and can hence be sampled from. We formalise this in Section \ref{sec:inf} as Theorem\,\ref{thm:1}. A further consequence of this result is Theorem \ref{thm:2} which demonstrates that sampling from the posterior distribution can be done efficiently.  Overall, this means that the Markov blanket of $p$ query nodes, can be estimated efficiently \textit{without explicitly inferring the entire network}.

An overview of our approach is presented in Fig.\,\ref{blockwise}, where the matrix $\ma{W}$ is partitioned similarly to \cite{wang2012bayesian}. More precisely, we consider the case where $p > 1$. The difference in the shading of the blocks in $\ma{W}$ indicates that estimation of $\ma{W}_{11}$ and $\ma{W}_{12}$ (and hence $\ma{W}_{21}$) is invariant of estimating $\ma{W}_{22}$. The core results described in the previous paragraph are summarised. 

The remainder of this paper is structured as follows. We begin by exploring the block factorization of the Wishart likelihood in Section~\ref{sec:cgl}. We subsequently derive the posterior distribution and construct a Gibbs sampler to efficiently sample from the different blocks in Section~\ref{sec:inf}. Section~\ref{sec:copula} describes how Bayesian Markov blanket estimation can be extended to deal with mixed data types with the copula framework. Finally, we demonstrate the practical applicability of the scheme in Section~\ref{sec:exp} with examples of artificial and real data.

\section{Likelihood Model}
\label{sec:cgl}

\paragraph{Problem Formulation}
Assume $\ma{X} \in \mathbb{R}^{(p+q)\times n}$ is a given matrix with $n$ independent observations. We are interested in estimating the Markov blanket of $p$ query variables with respect to the $q$ remaining variables in the data matrix. The sample covariance $\ma{S}$ follows the Wishart distribution $\ma{S} \sim \mathcal{W}_{p+q} \left(n, \ma{\Sigma} \right)$ with $n$ degrees of freedom. That is, $p(\ma{S}) \propto \det{\ma{W}}^{\frac{n}{2}} \exp \trace \left( -\frac{1}{2} \ma{W} \ma{S} \right)$. Assume that $\ma{S}$ and $\ma{W}$ are partitioned according to
\begin{equation}
\begin{aligned}
	\ma{S} = \bordermatrix{
		~ & p & q \\
		p & \ma{S}_{11} & \ma{S}_{12} \\
		q & \ma{S}_{21} & \ma{S}_{22}
	}, \quad \ma{W} = \bordermatrix{
		~ & p & q \\
		p & \ma{W}_{11} & \ma{W}_{12} \\
		q & \ma{W}_{21} & \ma{W}_{22}
	},
	\label{eq:partitioning}
\end{aligned}
\end{equation}
where the matrices have been reordered such that the query variables lie in the upper left block. Given $\ma{S}$, we would like to infer $\ma{W}_{12}$, the Markov blanket of the $p$ variables that constitute the block $\ma{S}_{11}$. We restrict the problem to the case where $p \ll q$ such that $\ma{S}_{11}$ is small, corresponding to the few variables of interest, and $\ma{S}_{22}$ is large.

\paragraph{Independence structure} We begin by showing a blockwise independence structure of the likelihood, which builds the foundation of our model.
\begin{lemma}
\label{lemma:1} 
Let $\ma{W}_{22.1} = \ma{W}_{22} - \ma{W}_{21}\ma{W}_{11}^{-1}\ma{W}_{12}$ be the Schur complement of the block $\ma{W}_{11}$. Then ($\ma{W}_{11}, \ma{W}_{12}) \bot \ma{W}_{22\cdot1}|\ma{S}$, where $\bot$ denotes conditional independence. That is,
\begin{equation}
   p(\ma{W}_{11}, \ma{W}_{12}, \ma{W}_{22\cdot1}|\ma{S}) = p(\ma{W}_{11}, \ma{W}_{12}|\ma{S})p(\ma{W}_{22\cdot1}|\ma{S}).
\label{eq:blockwiseindep}
\end{equation}
\end{lemma}
The proof of this lemma can be found in the supplementary document, and is analogous to Gupta and Nagar~[Chapter 3, pp. 94--95]\cite{gupta1999matrix}. A consequence of the conditional independence of the blocks $(\ma{W}_{11}, \ma{W}_{12})$ and $\ma{W}_{22\cdot1}$ in Lemma~\ref{lemma:1} is that we can derive the posterior distribution of the $\ma{W}_{11}$ and $\ma{W}_{12}$ blocks, and subsequently sample from them to deduce the Markov blanket. We discuss this in detail in the next section.

\section{Posterior Inference}
\label{sec:inf}
In this section we define the prior of our model and note that it also possesses the fundamental blockwise independence structure proved in Lemma~\ref{lemma:1} for the likelihood. The posterior distribution of $\ma{W}_{11} | \ma{W}_{12}, \ma{S}$ and $\ma{W}_{12} | \ma{W}_{11}, \ma{S}$ is therefore independent of $\ma{W}_{22}$. We derive this posterior distribution and show that it has an analytical form, which is formulated as Theorem~\ref{thm:1}. We subsequently demonstrate how to efficiently sample from it in Section~\ref{subs:sampling}.

\subsection{The Posterior Distribution}
\label{subs:posterior}
The natural conjugate prior to our likelihood is the Wishart distribution. However, in order to ensure sparsity, we also use a double exponential prior as in Wang~\cite{wang2012bayesian}. Since our focus is on the Markov blanket, we only place the latter on the block $\ma{W}_{12}$. This results in a compound prior:
\begin{equation}
\begin{aligned}
		p(\ma{W}) &= \mathcal{W}(p+q+1, \mai) p \left( \ma{W}_{12} | \ma{T} \right) p \left( \ma{T} | \gamma \right) \\
		&\propto \exp \trace \left( -\tfrac{1}{2} \mai \ma{W} \right) \prod_{w_{ij} \in \ma{W}_{12}} \tfrac{1}{\sqrt{2 \pi t_{ij}}} \exp \left( -\frac{w_{ij}^2}{2 t_{ij}} \right) \tfrac{\gamma^2}{2} \exp \left( -\tfrac{\gamma^2}{2} t_{ij} \right),
\end{aligned}
\label{eq:prior}
\end{equation}
where $\ma{T} = \{t_{ij}\}$ are inverse-Gaussian distributed scale parameters introduced by Wang~\cite{wang2012bayesian}:
\begin{equation}
	t_{ij}^{-1} \sim \mathcal{IG} \left( \sqrt{\gamma^2 / w_{ij}^2}, \gamma^2 \right)
\label{eq:t}
\end{equation}
and $\gamma$ is a hyperparameter. Most importantly, the compound prior does not affect the independence structure proved for the likelihood in Lemma~\ref{lemma:1}. Multiplying the prior introduced in Eq.~(\ref{eq:prior}) by the likelihood yields the posterior distributions for blocks $\ma{W}_{12}$ and $\ma{W}_{11}$.


We now state the main result of this paper. Specifically, we show that the posterior distribution required to estimate the Markov blanket can be expressed in an analytical form.  

Let the Matrix Generalised Inverse Gaussian (MGIG) distribution \cite{butler1998generalized} be defined by probability density function with parameter $\lambda$:
\begin{equation}
\begin{aligned}
    p(\ma{M}; \lambda, \ma{A}, \ma{B}) &\propto \det(\ma{M})^{-\lambda-1} \exp \trace \left( -\frac{1}{2} ( \ma{A} \ma{M} + \ma{B} \ma{M}^{-1} ) \right).
\end{aligned}
\label{eq:mgig}
\end{equation}

\begin{thm}
\label{thm:1}
The posterior conditionals $\ma{W}_{12}|\ma{W}_{11},\ma{S}, \ma{T}$ and $\ma{W}_{11}|\ma{W}_{12},\ma{S}, \ma{T}$ admit an analytical form:
\begin{thmenum}
\item \label{thm:1:1} 
Vectorised rows of $\ma{W}_{12}$ follow a joint normal distribution
\begin{equation}
    vec(\ma{W}_{12}^\T) | \ma{W}_{11}, \ma{S}, \ma{T} \sim \mathcal{N}_{pq} \left( vec(-(\ma{S}_{22}+\mai)^{-\T} \ma{S}_{12}^\T \ma{W}_{11}^\T), \ma{C}^{-1} \right),
\label{eq:postNormal}
\end{equation}
where $\ma{C} = \ma{W}_{11}^{-1} \otimes (\ma{S}_{22} + \ma{I}) + \diag \left(\ma{D}_1, \ldots, \ma{D}_p \right)$ be the covariance matrix, and $\ma{D}_i = \diag \left( (T_{i \cdot})^{-1} \right)$ be diagonal matrices containing $T_{i\cdot} = (t_{i1}, \ldots, t_{iq})$.
\item \label{thm:1:2} $\ma{W}_{11}$ follows the Matrix Generalised Inverse Gaussian (MGIG) distribution:
\begin{equation}
    \ma{W}_{11} | \ma{W}_{12}, \ma{S}, \ma{T} \sim \mathcal{MGIG}_{p \times p} \left( \frac{n}{2}+p, \ma{W}_{12} (\ma{S}_{22} + \mai) \ma{W}_{21}, \ma{S}_{11} + \mai\right).
\label{eq:mgigdist}
\end{equation}
\end{thmenum}
\end{thm}

\begin{proofsketch}
The posterior maintains the conditional independence structure proved for the likelihood in Lemma~\ref{lemma:1}, i.e.\ $p(\ma{W}_{11}, \ma{W}_{12}, \ma{W}_{22\cdot1}|\ma{S}, \ma{T}) = p(\ma{W}_{11}, \ma{W}_{12}|\ma{S}, \ma{T})p(\ma{W}_{22\cdot1}|\ma{S}, \ma{T})$, because the compound prior defined in Eq.~(\ref{eq:prior}) is element-wise independent. Derivations of the distributions in Eqs.~(\ref{eq:postNormal}) and (\ref{eq:mgigdist}) follow from factorising the posterior and rearranging terms. Relevant calculations are provided in the supplementary document.
\end{proofsketch}

Theorem~\ref{thm:1} shows that estimation of the Markov blanket of the $p$ query variables only requires sampling from the posterior conditionals of $\ma{W}_{11}$ and $\ma{W}_{12}$, which both have an analytical form while remaining independent of $\ma{W}_{22}$. 
Therefore, the amount of parameters in the Markov blanket that need to be estimated, scales linearly with $q$. This is an improvement over the Bayesian graphical lasso~\cite{wang2012bayesian} approach, where this number grows quadratically with $q$.
Theorem~\ref{thm:1} also provides us with the particular distributions to sample from. Next, we demonstrate how this sampling can be done efficiently.

\begin{algorithm}[t]
    \KwIn{Sample covariance matrix $\ma{S}$}
    \KwOut{Markov blanket $\ma{W}_{12}$}
    \While{not converged} {
        $T_{ij} \quad \sim \quad \mathcal{IG} \left( \sqrt{\gamma^2 / w^2_{ij}}, \gamma^2 \right)$ \\
        $vec(\ma{W}_{12}^\T) | \ma{W}_{11}, \ma{S} \quad \sim \quad \mathcal{N}_{pq} \left( vec(-(\ma{S}_{22}+\mai)^{-\T} \ma{S}_{12}^\T \ma{W}_{11}^\T), \ma{C}^{-1} \right)$ \\
        $\ma{W}_{11} | \ma{W}_{12}, \ma{S} \quad \sim \quad \mathcal{MGIG}_{p \times p} \left( -\tfrac{1}{2}(n+p+1), \ma{W}_{12} ( \ma{S}_{22} + \mai) \ma{W}_{21}, \ma{S}_{11} + \mai \right)$ \\
    }
\caption{Block Gibbs sampling scheme for the posterior.}
\label{alg:gibbs}
\end{algorithm}

\subsection{Efficiency of Sampling from the Posterior}
\label{subs:sampling}
The blockwise Gibbs sampling scheme for estimating the Markov blanket is summarised in Algorithm~\ref{alg:gibbs}. This sampling scheme consists of iterative resampling of $\ma{W}_{12} | \ma{W}_{11}, \ma{S}, \ma{T}$ and of $\ma{W}_{11} | \ma{W}_{12}, \ma{S}, \ma{T}$, according to their definitions in Theorem~\ref{thm:1}.

The distribution of $\ma{W}_{12} | \ma{W}_{11}, \ma{S}, \ma{T}$ is given by Theorem~\ref{thm:1:1}. The vectorised rows of $\ma{W}_{12}| \ma{W}_{11}, \ma{S}, \ma{T}$ follow a joint normal distribution. For $\ve{v} = vec(\ma{S}_{12}^\T)$, the distribution further simplifies to
\begin{equation}
	vec(\ma{W}_{12}^\T) | \ma{W}_{11}, \ma{S} \sim \mathcal{N}_{pq} \left(-\ma{C}^{-1} \ve{v}, \ma{C}^{-1} \right) .
\label{Cinversion}
\end{equation}
The majority of the computational cost incurred in our method arises from sampling from this joint normal distribution. Eq.~(\ref{Cinversion}) requires us to invert $\ma{C}$, which is of size ${pq \times pq}$. Note that $\ma{C}$ cannot be represented as a covariance tensor of a matrix normal distribution. Therefore, na\"ive inversion of $\ma{C}$ using a standard Cholesky decomposition would cost $\mathcal{O}(p^3q^3)$ operations. Our efficient sampling strategy exploits the structure of this matrix, which is the foundation of Theorem~\ref{thm:2}.

\begin{thm}
\label{thm:2}
    Sampling from the distribution in Theorem~\ref{thm:1:1} requires $\mathcal{O}(pq^3)$ operations.
\end{thm}

\begin{proofsketch}
We expand the Kronecker product of matrix $\ma{C} \in \mathbb{R}^{pq \times pq}$, which comprises $p$ blocks of size $q \times q$:
\begin{equation}
	\ma{C} = \begin{pmatrix}
			u_{11} (\ma{S}_{22} + \mai) + \ma{D}_1 & u_{12} (\ma{S}_{22} + \mai) & \cdots \\
			u_{21} (\ma{S}_{22} + \mai) & \ddots & \\
			\vdots & \cdots & u_{pp} (\ma{S}_{22} + \mai) + \ma{D}_p
	\end{pmatrix}
\end{equation}
where $\ma{U} = \ma{W}_{11}^{-1}$ is the inverted upper diagonal block. We observe a regular structure within the blocks in $\ma{C}$: Matrices $\ma{D}_i$ are added to the diagonals blocks only, and the non-diagonal blocks only differ by scalar factors $u_{ij}$. With a blockwise Cholesky factorisation, the inversion requires only $pq^3$ operations. Since the Cholesky decomposition of the blocks also only differs by a factor, we can store its intermediate result. 
\end{proofsketch}


Theorem~\ref{thm:1:2} states that $\ma{W}_{11}|\ma{W}_{12}, \ma{S}, \ma{T}$ follows the MGIG distribution. In order to sample from this distribution, we make use of a result by Bernadac~\cite{bernadac1995random}. It introduces a representation of an MGIG-distributed random variable as a limit of a random continued fraction of Wishart-distributed random variables. The interested reader should refer to \cite{letac2000symmetric,bernadac1995random,koudou2014characterizations} for the details. Drawing samples from the MGIG thus reduces to iterated sampling from the Wishart distribution. In practice, we observe the convergence of the random continued fraction within few iterations. The complexity of sampling from the distribution derived in Theorem~\ref{thm:1:2} does not depend on $q$.

\section{Extension to Gaussian Copula}
\label{sec:copula}

We extend the model for non-Gaussian and mixed continuous/discrete data by embedding it within a copula construction.
Copulas describe the dependency in a $r$-dimensional joint distribution $F(Y_1, \ldots, Y_r)$ and represent an invariance class with respect to the marginal cumulative distribution functions (cdf) $F_i$. In our model, $r = p+q$.
For continuous cdfs, Sklar's theorem \cite{sklar1959fonctions} guarantees the existence and uniqueness of a copula $C$, such that $F(Y_1, \ldots, Y_r) = C\left( F_1(Y_1), \ldots, F_r(Y_r) \right)$. For discrete cdfs, this leads to an identifiability problem \cite{genest2007primer}, such that established methods on empirical marginals \cite{liu2009nonparanormal} cannot be used anymore, but a valid copula can still be constructed \cite{genest2007primer}.
For our purpose, we follow the semi-parametric approach by \cite{hoff2007extending} and restrict our model to the parametric Gaussian copula, but we do not restrict the data to be Gaussian and treat them in a non-parametric fashion. The Gaussian copula inherently implies latent variables $X_i = \Phi^{-1}(F_i(Y_i))$. Our model under consideration is
\begin{equation}
    (X_1, \ldots, X_r)^\T \sim \mathcal{N}_r \left( \vez, \ma{\Sigma} \right), \quad Y_i = F_i^{-1} \left( \Phi( X_i) \right)
\end{equation}
where $F_i^{-1}$ denotes the $i$th generalised inverse of continuous or discrete cdfs, $\ma{X}$ are the latent variables, and $\ma{Y}$ are the observations.

Following \cite{hoff2007extending}, inference in the latent variables uses the non-decreasing property of discrete cdfs for transforming the observed variables to the latent space. This guarantees that for observations $y_{ik} < y_{il}$ we also have $x_{ik} < x_{il}$, and more generally, $\ma{X}$ must lie in the set
\begin{equation}
    \mathcal{D} = \{ \ma{X} \in \mathbb{R}^{r \times n} : \max(x_{ik}: y_{ik} < y_{ij} ) < x_{i, j} < \min( x_{ik}: y_{ij} < y_{ik} ) \}
\end{equation}
The data likelihood can then be written as
\begin{equation}
\begin{aligned}
    p(\ma{Y} | \ma{\Sigma}, F_1, \ldots, F_r) &= p(\ma{X} \in \mathcal{D}, \ma{Y} | \ma{\Sigma}, F_1, \ldots, F_r) \\
    &= p(\ma{X} \in \mathcal{D} | \ma{\Sigma}) p(\ma{Y} | \ma{X} \in \mathcal{D}, \ma{\Sigma}, F_1, \ldots, F_r)
\end{aligned}
\end{equation}
and estimation of $\ma{\Sigma}$ is performed on maximising the sufficient statistics $p(\ma{X} \in \mathcal{D} | \ma{\Sigma})$ only, thus treating the marginals $F_i$ as nuisance parameters. Bayesian inference for $\ma{\Sigma}$ is achieved by a Markov chain having stationary distribution at the posterior $p(\ma{\Sigma} | \ma{X} \in \mathcal{D}) \propto p(\ma{\Sigma}) p(\ma{X} \in \mathcal{D} | \ma{\Sigma})$, where a inverse-Wishart prior $p(\ma{\Sigma})$ is used. Posterior inference can be achieved with a Gibbs sampler, which draws alternately between $\ma{X} | \ma{\Sigma}, \ma{Y}$ and $\ma{\Sigma} | \ma{X}$. This sampler extends Alg.~\ref{alg:gibbs} with an additional outer loop for inferring the latent variables. The Markov blanket is then iteratively estimated on these variables. The sampling scheme easily accommodates for missing values, when omitting conditioning on the set $\mathcal{D}$.

The presented framework is very useful in practice, since the invariance class of copulas extend the model to non-Gaussian data. With the additional stochastic transformation to the latent space, we can use discrete variables and allow missing values. In real world applications, it becomes apparent that this is a very valuable extension.

\section{Experiments}
\label{sec:exp}

\subsection{Artificial Data}
As a first experiment, we attempt to highlight the differences in inference between the Bayesian Markov blanket (BMB) and Bayesian Graphical Lasso (BGL) procedures. We construct an artificial network with $100$ variables, where our interest is confined to only the Markov blanket between $p = 10$ query variables and the $q = 90$ remaining variables. In order to create networks with a ``small-world'' flavour containing \emph{hubs}, i.e. nodes with very high degree, the connectivity structure of the inverse covariance matrix $\ma{W}$ is generated by a beta-binomial model. Edge weights are sampled uniformly from the interval $[0.3,1]$, and  edge signs are randomly flipped. Finally, positive definiteness is guaranteed by adding a suitable constant (related to the smallest eigenvalue) to the diagonal.    
 This process produces a sparse network structure where the majority of edges are connected to only a few single nodes. Note that many real-world networks exhibit such small-world properties. 

\begin{figure}[!htbp]
  \centering
  \includegraphics[width=\textwidth]{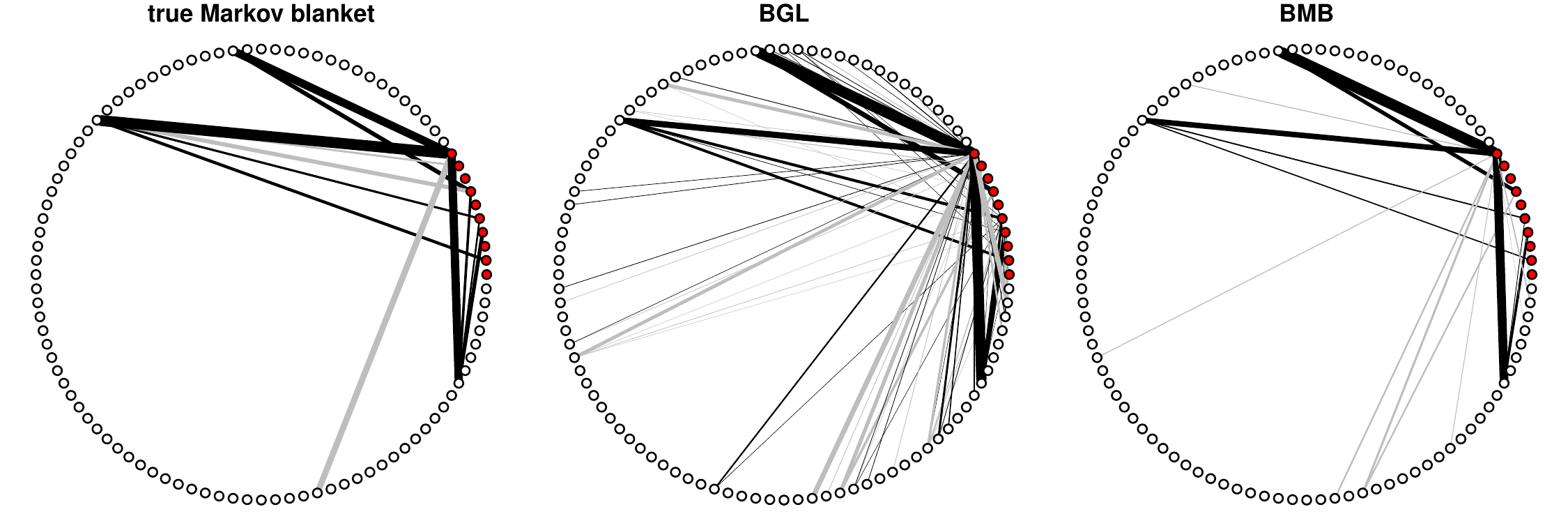}
  \caption{One exemplary Markov blanket ($p = 10$, $q = 90$) and its reconstruction by BGL and BMB. Note that the graphs \emph{only} display edges between $p$ query and $q$ remaining variables. Red nodes represent query variables,  white nodes represent all other variables. Black and grey edges correspond to positive and negative edge signs, respectively.}
  \label{artificial_experiment_example_networks}
\end{figure}

\begin{figure}[!htbp]
  \centering
  \includegraphics[width=0.7\textwidth]{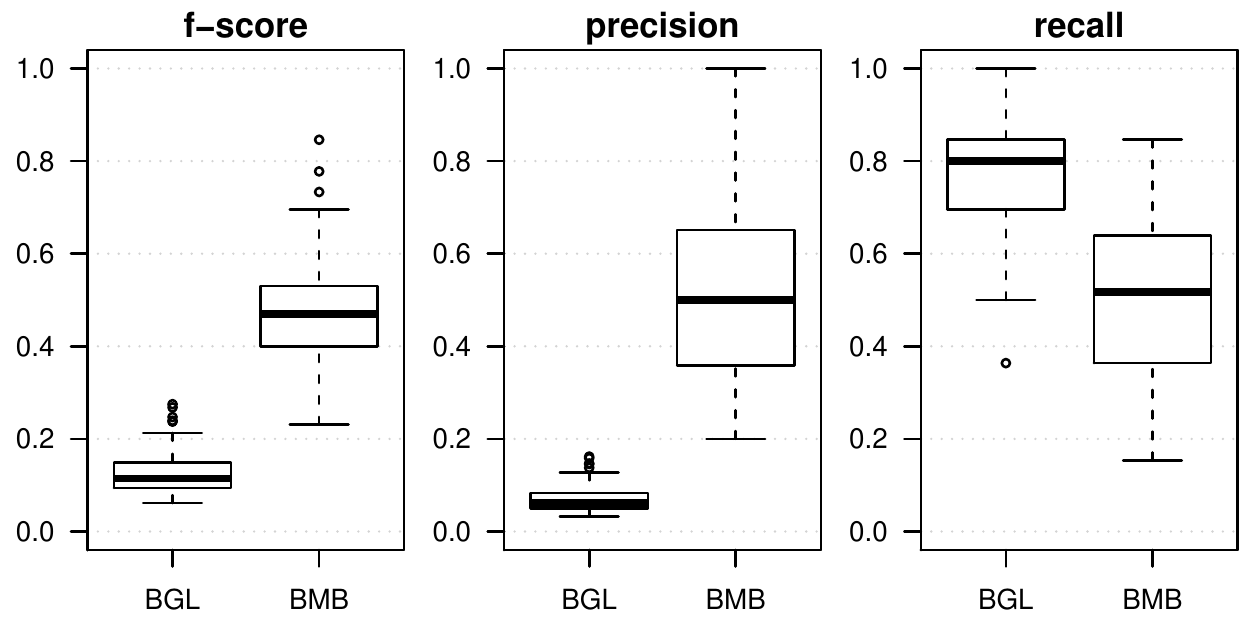}
  \caption{Performance of inferred Markov blankets from $100$ datasets.}
  \label{artificial_experiment_fscore}
\end{figure}

Next, we draw $n = 1000$ independent samples from a zero-mean normal distribution with covariance matrix $\ma{W}^{-1}$ and compute the sample covariance $\ma{S}$. Fig.~\ref{artificial_experiment_example_networks} depicts a true Markov blanket and its reconstruction by BGL and BMB using the same sparsity parameter $\lambda = 200$. Both methods were run side-by-side for $700$ MCMC samples after an initial burn-in phase of $300$ samples. From the sampled networks, a representative network structure is constructed by thresholding  based on a $85\%$ credibility interval. 
We repeat the above procedure to obtain a total of $100$ datasets. The quality of reconstructed networks is measured in terms of $f$-score (harmonic mean of precision and recall) between the true and inferred Markov blanket. When computing precision and recall, inferred edges with edge weights having the wrong sign are counted as missing. Both models share the same sparsity parameter $\lambda$, which in this experiment was selected such that for BMB recall and precision have roughly the same value. The results are depicted as box plots in Fig.~\ref{artificial_experiment_fscore}, from which we conclude that there are indeed substantial differences in both models. In particular, BGL has the tendency to introduce many unnecessary edges in comparison to BMB. As a result, BGL achieves high recall and low $f$-score. Since both methods are based on the same likelihood model and (almost) the same prior, the observed differences can only be attributed to differences in the inference procedure: BGL infers a network by iterating over \emph{all} variables and their neighbourhood systems, whereas BMB only estimates the elements in  $\ma{W}_{11}$ and $\ma{W}_{12}$. 

\begin{figure}[!htbp]
  \centering
  \includegraphics[width=\textwidth]{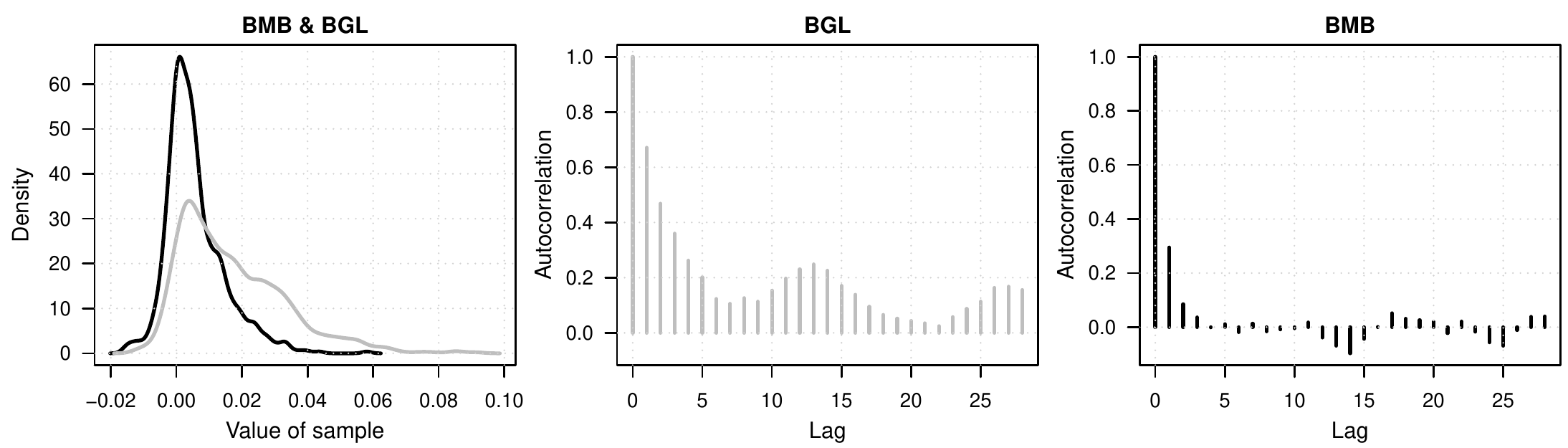}
  \caption{Density and auto-correlation of the Markov chain for a single variable in the Markov blanket. Gray refers to BGL, black to BMB.}
  \label{artificial_experiment_mixing}
\end{figure}

To further study the influence of the different Gibbs sampling strategies, we examine tracer plots and auto-correlations of individual variables  in Fig.~\ref{artificial_experiment_mixing}. In almost all cases, BGL shows significantly higher auto-correlation and poor convergence. In contrast, Markov chains in the BMB sampler seems to mix much better, typically leading to posteriors with smaller bias and variance. While only one example is shown in the figure, similar results can be seen for basically all variables in the network. 
Further, we experience a substantial decrease in run-time, even for these relatively small networks: computing $1000$ MCMC samples for BMB finished on average after $100$ seconds, while BGL typically consumed around $370$ seconds. Since BGL requires an additional sampling loop over \emph{all} variables, datasets with large $\ma{S}_{22}$ quickly become problematic for BGL. We further explore these differences in the next section for a large real-world application.

\subsection{Real Data}
\label{sec: real}
To demonstrate the practical significance of Markov blanket estimation, we turn to the analysis of \emph{colorectal cancer}, which in 2012 ranked among the three most common types of cancer globally \cite{worldcancerreport2014}. The data set introduced in \cite{sheffer2009} is publicly available and contains gene expression measurements from biopsies of $n = 260$ cancer patients. A separate table captures discrete/categorical clinical traits such as sex, age or pathological staging/grading.  In this context, one particularly interesting research question is to identify connections between the $p$ (macroscopic) clinical descriptors and the $q$ (molecular) gene expression measurements. 

\begin{figure}[!htbp]
  \centering
  \includegraphics[width=0.49\textwidth]{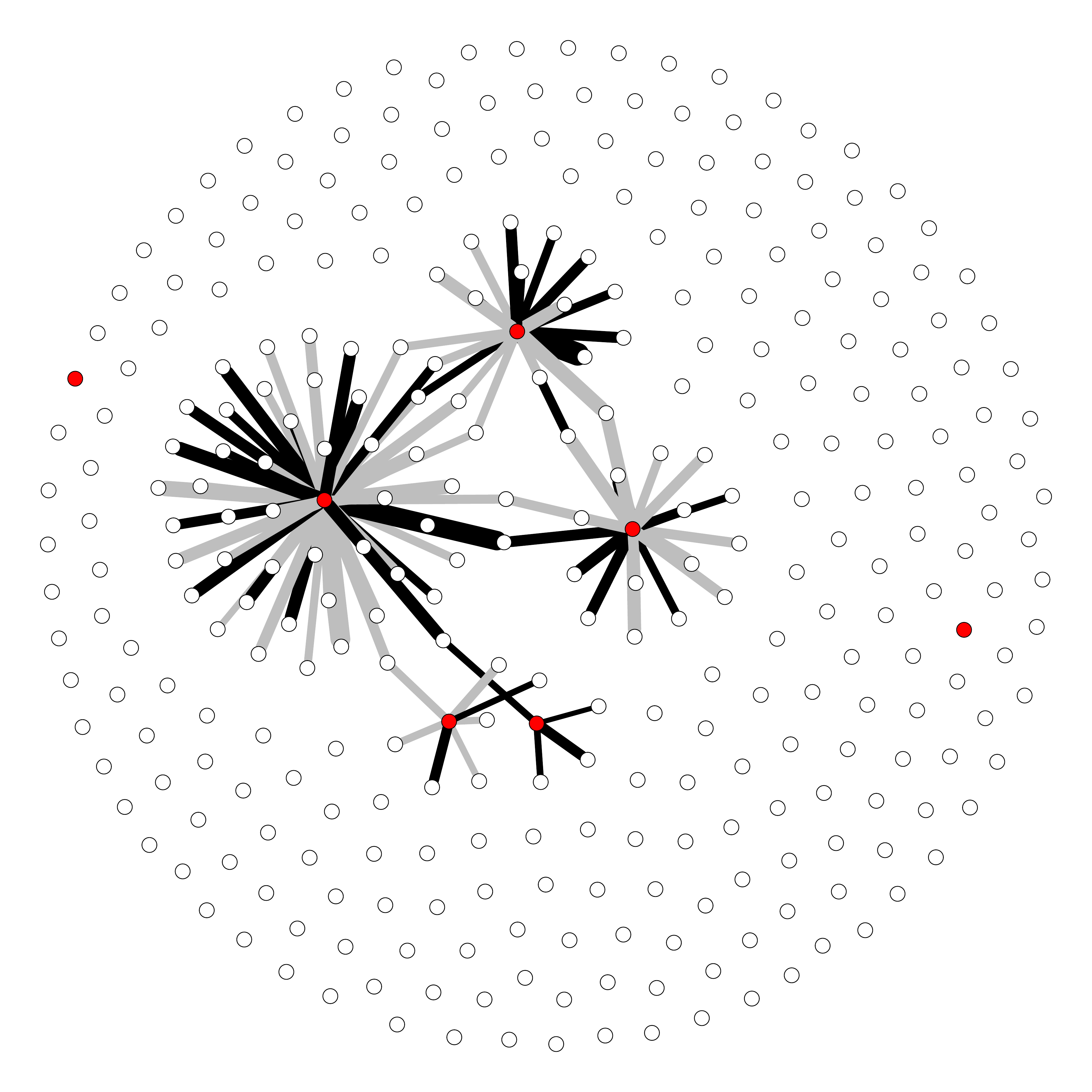}
  \includegraphics[width=0.49\textwidth, trim=13mm 14mm 14mm 14mm, clip]{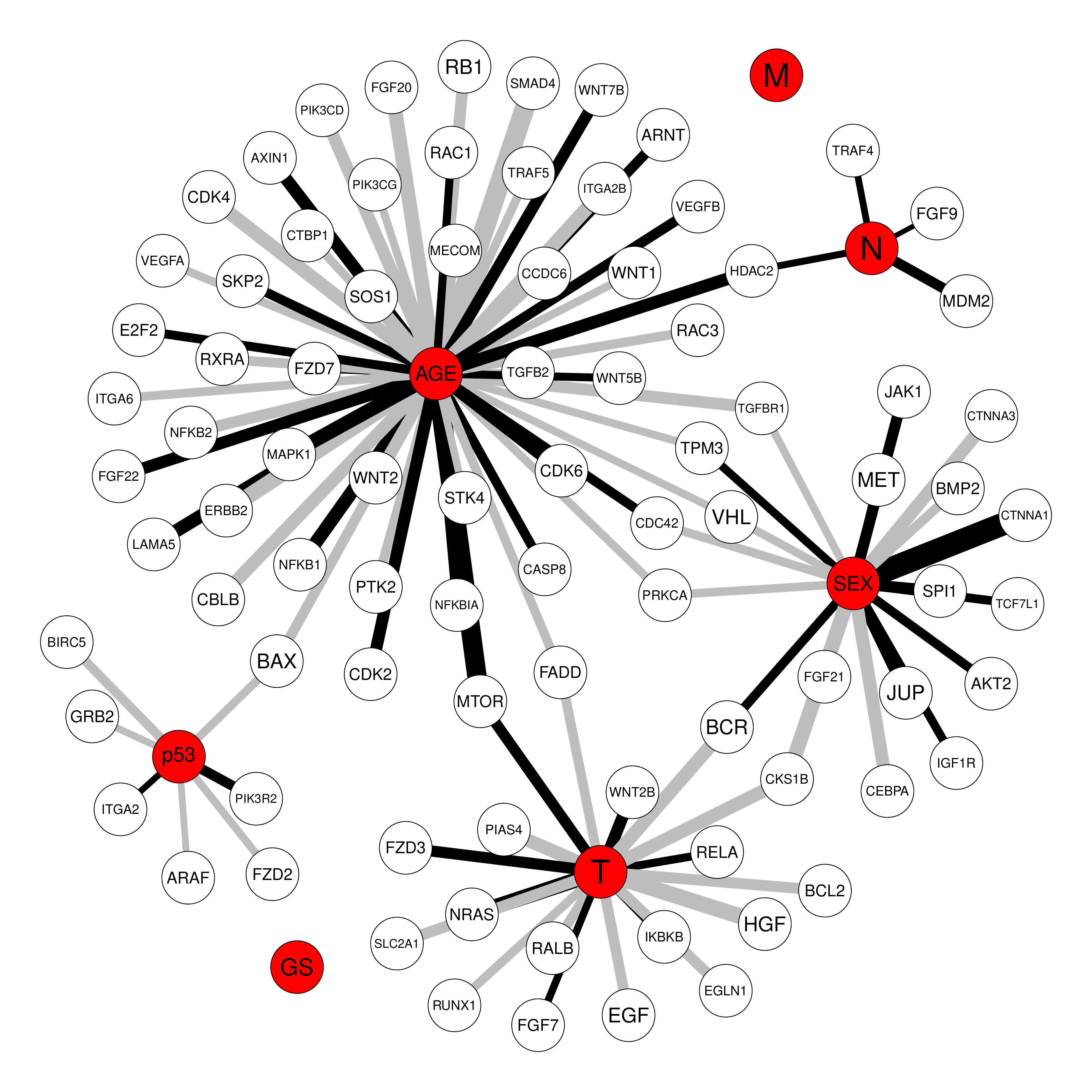}
  \caption{Sparse Markov blanket between $p = 7$ clinical features (red nodes) and $q = 312$ genes in colorectal cancer \cite{sheffer2009}. Overview of all variables/nodes (left) and enlarged, fully labeled subgraph (right).}
  \label{real_world_markov_blanket}
\end{figure}

Among the $13\,400$ genes contained in the dataset, we focus on a specific subset, the so-called ``\emph{Pathways in cancer}" as defined in the KEGG database\footnote{Kyoto Encyclopedia of Genes and Genomes, \url{http://www.genome.jp/kegg/pathway.html}}. This particular subset comprises a general class of genes which are known to be involved in various biological processes linked to cancer. For this experiment, we have $q = 312$ candidate genes and $p = 7$ query variables. These are the age and sex of the patient as well as the \emph{TNM} classification, cancer group stage (\emph{GS}) and mutation of the tumor suppressor protein \emph{p53}. Since the observations have mixed continuous/discrete data types with missing values, the Markov blanket estimation is extended by a semi-parametric Gaussian copula framework \cite{hoff2007extending}. Based on this, we calculate $5\,000$ MCMC samples, 
which finally leads to the Markov blanket in Fig.~\ref{real_world_markov_blanket}. 

The resulting network structure confirms some well-known properties like the confounding effect of the age and sex variables, both of which (correctly) link to a large number of genes. For example, \emph{FGF21} exhibits significant differences in male and female subjects \cite{fgf21_sex}, and \emph{CTNNA1} shares connections to survival time in men \cite{catenin_sex}. Similarly, \emph{mTOR}, the \emph{mechanistic target of rapamycin}, not only represents a key element for cell signaling that triggers a cascade of immune-related pathways, but its function also depends heavily on a subject's age \cite{mtor_age}. 
Further, we are able to identify a very interesting network structure around the variable \emph{tumor size} \emph{T}: almost all direct neighbors control either cell growth (\emph{EGLN1} \cite{egln1_t}, \emph{RELA} \cite{rela_t}, \emph{HGF} \cite{hgf_t,hgf_t_2} and others) or cell death (\emph{BCL2}, \emph{FADD}). Cancer typically affects the balance between these two fundamental processes and their deregulation eventually leads to tumor development. A second subgraph concerns variable \emph{N}, the degree of spread to regional lymph nodes, which is expressed in $4$ levels $N0$ to $N3$. Here, all genes in the neighbourhood correspond to the lymphatic system and its direct responses to malignant cell growth, which was confirmed for \emph{FGF9} \cite{fgf9_n}, \emph{MDM2} \cite{mdm2_n,mdm2_n_2} and \emph{TRAF4} \cite{traf4_n} among others. 

Despite the study's focus on colorectal cancer and specifics of the intestinal system, the inferred Markov blanket is able to explain rather general properties in accordance with findings in the medical literature. Altogether, this nicely illustrates how the Gaussian copula framework complements the Bayesian Markov blanket estimation -- especially pertaining to the clinical domain with mixed observations and missing values.

In contrast to our approach, the high dimensionality of this dataset imposes severe problems for BGL. For BMB, 5000 Gibbs sweeps could be computed in 2 hours, and MCMC diagnosis did not show any severe convergence problems. For BGL, however, the same number of iterations already took $122$ hours ($\approx 5$ days), and we observed similar (and sometimes severe) mixing problems as described in the previous section. 

\section{Conclusion}
We have presented a Bayesian perspective for estimating the Markov blanket of a set of query nodes in an undirected network. In our experience, it is often the case that we estimate a full network but interpret only part of it. This is especially true in a context where portions of our data are qualitatively different. Here, we would be more interested in establishing the links between these portions, rather than examining the links within the portions themselves. Markov blanket estimation is hence an interesting and relevant sub-problem of network estimation, particularly in high dimensional settings. Existing methods such as the Bayesian graphical lasso iterate through the individual variables to estimate an entire network. While there are several situations in which inference of the entire network is required, there are also cases in which we are only interested in the neighbourhood of a small subset of query variables; for these instances, iterating through all the variables is unnecessary. 

In this paper, we explored the blockwise factorisation of the Wishart likelihood in combination with a suitable choice of prior. Our primary contribution in Theorem~\ref{thm:1} shows that the resulting posterior distribution of the Markov blanket of a set of query nodes has an analytic form, and is independent of a large portion of the network. The analytic form allows us to explore potentially large neighbourhoods where the Bayesian graphical lasso reaches its limits.
We also demonstrated that sampling from the posterior of the Markov blanket is more efficient than the Bayesian graphical lasso. Moreover, we observed fast convergence and superior mixing properties of the Markov chain. We attribute this to the improved flexibility of our sampling strategy.

Including a copula construct in the model further enhances its real world applicability, where mixed data and missing values are prevalent. A particular application in a medical setting is the colorectal example we considered in Section \ref{sec: real}. Using this approach allowed us to make interesting observations about the interactions between various clinical and genetic factors. Such insights could ultimately contribute to a better understanding of the disease.

\newpage
\appendix

\section{Model}

Let $\ma{S} \sim \mathcal{W}_{p+q} \left( n, \ma{W} \right)$ be Wishart distributed with $n$ degrees of freedom and inverse covariance $\ma{W} = \ma{\Sigma}^{-1}$, and define the following partitioning:
\begin{equation}
\begin{aligned}
	\ma{S} &= \bordermatrix{
		~ & p & q \\
		p & \ma{S}_{11} & \ma{S}_{12} \\
		q & \ma{S}_{21} & \ma{S}_{22}
	}, \quad \ma{W} = \bordermatrix{
		~ & p & q \\
		p & \ma{W}_{11} & \ma{W}_{12} \\
		q & \ma{W}_{21} & \ma{W}_{22}
	} .
\end{aligned}
\label{eq:likelihood}
\end{equation}

\paragraph{Likelihood}

\begin{equation}
\begin{aligned}
    p(\ma{S} | \ma{W}) &\propto \det(\ma{W})^\frac{n}{2} \exp \trace \left( -\frac{1}{2}	\ma{W} \ma{S} \right) \\
    &= \det(\ma{W}_{11})^\frac{n}{2} \det(\ma{W}_{22 \cdot 1})^\frac{n}{2} \\
    &\quad \times \exp \trace \left( -\frac{1}{2} \left( \ma{W}_{11} \ma{S}_{11} + \ma{W}_{12} \ma{S}_{21} + \ma{W}_{21} \ma{S}_{12} + \ma{W}_{11}^{-1} \ma{W}_{12} \ma{S}_{22} \ma{W}_{21} \right) \right)\\
    &\quad \times \exp \trace \left( -\frac{1}{2} \ma{W}_{22 \cdot 1} \ma{S}_{22} \right)
\end{aligned}
\label{eq:sup:prior}
\end{equation}

\paragraph{Prior}

\begin{equation}
\begin{aligned}
		p(\ma{W}) &\propto \exp \trace \left( -\frac{1}{2} \mai \ma{W} \right) \prod_{w_{ij} \in \ma{W}_{12}} \frac{1}{\sqrt{2 \pi t_{ij}}} \exp \left( -\frac{w_{ij}^2}{2 t_{ij}} \right) \frac{\gamma^2}{2} \exp \left( -\frac{\gamma^2}{2} t_{ij} \right) \\
		&= \exp \trace \left( -\frac{1}{2} \mai \ma{W} \right) \prod_{w_{ij} \in \ma{W}_{12}} \frac{1}{\sqrt{2 \pi t_{ij}}} \prod_{i=1}^p \exp \left( -\frac{1}{2} \ve{\beta}_i^\T \ma{D}_i \ve{\beta}_i \right) \frac{\gamma^2}{2} \exp \left( -\frac{\gamma^2}{2} t_{ij} \right)
\end{aligned}
\end{equation}
Here, $\ve{\beta}_i = (\ma{W}_{12})_{i\cdot}$ denotes the $i$th row of $\ma{W}_{12}$, $\ma{D}_i = \diag \left( (T_{i \cdot})^{-1} \right)$, and $T_{i\cdot} = \diag \left(t_{i1}, \ldots, t_{ip} \right)$.

\paragraph{Posterior}

\begin{equation}
\begin{aligned}
    &p(\ma{W}_{11}, \ma{W}_{12}  | \ma{S}) \propto \det(\ma{W}_{11})^\frac{n}{2} \\
    &\quad \times \exp \trace \left( -\frac{1}{2} \left( \ma{W}_{11} (\ma{S}_{11} + \mai) + \ma{W}_{12} \ma{S}_{21} + \ma{W}_{21} \ma{S}_{12} + \ma{W}_{11}^{-1} \ma{W}_{12} (\ma{S}_{22} + \mai) \ma{W}_{21} \right) \right) \\
    &\quad \times \prod_{w_{ij} \in \ma{W}_{12}} \frac{1}{\sqrt{2 \pi t_{ij}}} \prod_{k=1}^p \exp \left( -\frac{1}{2} \ve{\beta}_k^\T \ma{D}_k \ve{\beta}_k \right) \frac{\gamma^2}{2} \exp \left( -\frac{\gamma^2}{2} t_{ij} \right)
\end{aligned}
\label{eq:posterior}
\end{equation}

\section{Wishart Distribution}

If $\ma{S} \sim \mathcal{W}_{p+q} \left(n, \ma{\Sigma} \right)$, $n > p+q-1$, then
\begin{equation}
\begin{aligned}
	p( \ma{S} ) &\propto \det(\ma{\Sigma})^{-\frac{n}{2}} \det(\ma{S})^{\frac{n-(p+q)-1}{2}} \exp \trace \left( -\frac{1}{2} \ma{\Sigma}^{-1} \ma{S} \right) \\
	&= \det(\ma{W})^{\frac{n}{2}} \det(\ma{S})^{\frac{n-(p+q)-1}{2}} \exp \trace \left( -\frac{1}{2} \ma{W} \ma{S} \right)
\end{aligned}
\end{equation}
where $\ma{W} = \ma{\Sigma}^{-1}$.


We will prove Lemma 1 from the paper.

\paragraph{Lemma 1} If $\ma{S} \sim \mathcal{W}_{p+q} \left( n, \ma{\Sigma} \right)$, then
\begin{equation}
		(\ma{W}_{11}, \ma{W}_{12}) \bot \ma{W}_{22 \cdot 1} | \ma{S}
\end{equation}
\paragraph{Proof} The proof is similar to \cite{gupta1999matrix}. Factorising the Wishart density according to
\begin{equation}
\begin{aligned}
    \ma{W} \ma{S} &= \ma{W}_{11} \ma{S}_{11} + \ma{W}_{12} \ma{S}_{21} + \ma{W}_{21} \ma{S}_{12} + \ma{W}_{22} \ma{S}_{22} \\
    &= \ma{W}_{11} \ma{S}_{11} + \ma{W}_{12} \ma{S}_{21} + \ma{W}_{21} \ma{S}_{12} + (\ma{W}_{22} - \ma{W}_{21} \ma{W}_{11}^{-1} \ma{W}_{12} + \ma{W}_{21} \ma{W}_{11}^{-1} \ma{W}_{12}) \ma{S}_{22} \\
    &= \ma{W}_{11} \ma{S}_{11} + \ma{W}_{12} \ma{S}_{21} + \ma{W}_{21} \ma{S}_{12}+ \ma{W}_{21} \ma{W}_{11}^{-1} \ma{W}_{12} \ma{S}_{22} + \ma{W}_{22\cdot1} \ma{S}_{22}
\end{aligned}
\end{equation}
and
\begin{equation}
		\det(\ma{W})^{\frac{n}{2}} = \det(\ma{W}_{11})^{\frac{n}{2}} \det(\ma{W}_{22\cdot1})^{\frac{n}{2}}
\end{equation}
the independence follows from
\begin{equation}
		p(\ma{W} | \ma{S}) = p(\ma{W}_{11}, \ma{W}_{12} | \ma{S}) p(\ma{W}_{22\cdot1} | \ma{S})
\end{equation}
where
\begin{equation}
\begin{aligned}
		&p(\ma{W}_{11}, \ma{W}_{12} | \ma{S}) \\
		&\propto \det(\ma{W}_{11})^{\frac{n}{2}} \exp \trace \left( - \frac{1}{2} \left( \ma{W}_{11} \ma{S}_{11\cdot2} + \ma{W}_{12} \ma{S}_{21} + \ma{W}_{21} \ma{S}_{12}+ \ma{W}_{21} \ma{W}_{11}^{-1} \ma{W}_{12} \ma{S}_{22} \right) \right)
\end{aligned}
\end{equation}
and
\begin{equation}
		p(\ma{W}_{22\cdot1} | \ma{S}) \propto \det(\ma{W}_{22\cdot1})^{\frac{n}{2}} \exp \trace \left( -\frac{1}{2} \left( \ma{W}_{22\cdot1} \ma{S}_{22}  \right) \right) .
\end{equation}

\section{Matrix Normal ($\mathcal{MN}$) Distribution}

$\ma{X} \in \mathbb{R}^{p\times n}$ follows a Matrix Normal distribution, if
\begin{equation}
\begin{aligned}
		\ma{X} &\sim \mathcal{MN}_{p \times n} \left(\ma{M}, \ma{\Sigma}, \ma{\Omega} \right) \\
		&\propto \frac{1}{(2 \pi)^{pn/2}} \det(\ma{\Sigma})^{-\frac{p}{2}} \det(\ma{\Omega})^{-\frac{n}{2}} \exp \trace \left( -\frac{1}{2} \ma{\Sigma}^{-1} (\ma{X} - \ma{M}) \ma{\Omega}^{-1} (\ma{X} - \ma{M})^\T \right)
\end{aligned}
\end{equation}
where mean $\ma{M} \in \mathbb{R}^{p \times n}$, column covariance $\ma{\Sigma} \in \mathbb{R}^{p \times p}$, and row covariance $\ma{\Omega} \in \mathbb{R}^{n \times n}$.

\paragraph{Lemma 2} If $\ma{S} \sim \mathcal{W}_{p+q} \left( n, \ma{\Sigma} \right)$ is Wishart distributed, then
\begin{equation}
\begin{aligned}
    \ma{W}_{12} | \ma{W}_{11}, \ma{S} &\sim \mathcal{MN}_{p \times q} \left( - \ma{W}_{11} \ma{S}_{12} \ma{S}_{22}^{-1}, \ma{W}_{11}, \ma{S}_{22}^{-1} \right)
\end{aligned}
\label{eq:mninwishart}
\end{equation}
is matrix normal distributed.

\paragraph{Proof} The proof is similar to \cite{gupta1999matrix}. Factorising the Wishart density according to
\begin{equation}
\begin{aligned}
    \ma{W} \ma{S} &= \ma{W}_{11} \ma{S}_{11} + \ma{W}_{12} \ma{S}_{21} + \ma{W}_{21} \ma{S}_{12} + \ma{W}_{22} \ma{S}_{22} \\
    &= \ma{W}_{11} (\ma{S}_{11} - \ma{S}_{12} \ma{S}_{22}^{-1} \ma{S}_{21} + \ma{S}_{12} \ma{S}_{22}^{-1} \ma{S}_{21}) + \ma{W}_{12} \ma{S}_{21} + \ma{W}_{21} \ma{S}_{12} \\ & \quad + (\ma{W}_{22} - \ma{W}_{21} \ma{W}_{11}^{-1} \ma{W}_{12} + \ma{W}_{21} \ma{W}_{11}^{-1} \ma{W}_{12}) \ma{S}_{22} \\
    &= \ma{W}_{11} \ma{S}_{11\cdot2} + \ma{W}_{11} \ma{S}_{12} \ma{S}_{22}^{-1} \ma{S}_{21} + \ma{W}_{12} \ma{S}_{21} + \ma{W}_{21} \ma{S}_{12}+ \ma{W}_{21} \ma{W}_{11}^{-1} \ma{W}_{12} \ma{S}_{22} + \ma{W}_{22\cdot1} \ma{S}_{22} \\
    &= \underbrace{ \ma{W}_{11} \ma{S}_{11\cdot2} }_{ \mathcal{W} } + \underbrace{  \ma{S}_{22} \left(\ma{W}_{12} + \ma{W}_{11} \ma{S}_{12} \ma{S}_{22}^{-1} \right)^\T \ma{W}_{11}^{-1} \left( \ma{W}_{12} + \ma{W}_{11} \ma{S}_{12} \ma{S}_{22}^{-1} \right) }_{ \mathcal{MN} } + \underbrace{ \ma{W}_{22\cdot1} \ma{S}_{22} }_{ \mathcal{W} }
\end{aligned}
\label{eq:normalfactorization}
\end{equation}
where we changed variables $\ma{W}_{22\cdot1} = \ma{W}_{22} - \ma{W}_{21} \ma{W}_{11}^{-1} \ma{W}_{12}$ with Jacobian $1$ and integrated over $\ma{W}_{22\cdot1}$.
Factorising the determinants according to
\begin{equation}
\begin{aligned}
\det(\ma{W})^{\frac{n}{2}} &= \det(\ma{W}_{11})^{\frac{n}{2}} \det(\ma{W}_{22\cdot1})^{\frac{n}{2}} \\
    &= \det(\ma{W}_{11})^\frac{n+p}{2} \underbrace{ \det(\ma{W}_{11})^{-\frac{p}{2}} }_{\mathcal{MN}} \det(\ma{W}_{22\cdot1})^\frac{n}{2}
\end{aligned}
\label{eq:normaldeterminant}
\end{equation}
and
\begin{equation}
\begin{aligned}
\det(\ma{S})^{\frac{n-(p+q)-1}{2}} &=  \det \left(\ma{S}_{22} \right)^{\frac{n-p-q-1}{2}} \det \left( \ma{S}_{11\cdot2} \right)^{\frac{n-p-q-1}{2}} \\
    &= \det(\ma{S}_{22})^\frac{n-p-2q-1}{2} \underbrace{ \det(\ma{S}_{22}^{-1})^{-\frac{q}{2}} }_{\mathcal{MN}} \det( \ma{S}_{11\cdot2} )^\frac{n-p-q-1}{2}
\end{aligned}
\end{equation}
gives the desired result.

\section{Sampling from $vec(\ma{W}_{12}^\T) | \ma{W}_{11}, \ma{S}, \{t_{ij}\}$}

We will prove Theorem 1, part 1 from the paper.

\paragraph{Theorem 1, Part 1} Vectorised rows of $\ma{W}_{12}$ follow a joint Normal distribution
\begin{equation}
    vec(\ma{W}_{12}^\T) | \ma{W}_{11}, \ma{S} \sim \mathcal{N}_{pq} \left( vec(-(\ma{S}_{22}+\mai)^{-\T} \ma{S}_{12}^\T \ma{W}_{11}^\T), \ma{C}^{-1} \right)
\label{eq:sup:postNormal}
\end{equation}
where the matrix is $\ma{C} = \ma{W}_{11}^{-1} \otimes (\ma{S}_{22} + \mai) + \diag \left(\ma{D}_1, \ldots, \ma{D}_p \right)$, and $\ma{D}_i = \diag \left( (T_{i \cdot})^{-1} \right)$ is a diagonal matrix containing $T_{i\cdot} = (t_{i1}, \ldots, t_{iq})$.

\paragraph{Proof}

According to Lemma 2, the likelihood in Eq.~(\ref{eq:likelihood}) can be expressed as a Matrix Normal distribution as in Eq. (\ref{eq:mninwishart}). Including the Wishart part of the prior changes the distribution in Eq.~(\ref{eq:mninwishart}) to
\begin{equation}
    \ma{W}_{12} | \ma{W}_{11}, \ma{S} \sim \mathcal{MN}_{p \times q} \left( - \ma{W}_{11} \ma{S}_{12} (\ma{S}_{22}+\mai)^{-1}, \ma{W}_{11}, (\ma{S}_{22} + \mai)^{-1} \right)
\label{eq:augmentedding}
\end{equation}
which is equivalent to
\begin{equation}
    vec(\ma{W}_{12}^\T) | \ma{W}_{11}, \ma{S} \sim \mathcal{N}_{pq} \left( vec(- (\ma{S}_{22}+\mai)^{-\T} \ma{S}_{12}^\T  \ma{W}_{11}^\T ), \ma{W}_{11}^{-1} \otimes (\ma{S}_{22} + \mai) \right)
\label{eq:covtensor}
\end{equation}
where $vec(\ma{W}_{12}^\T)$ are the vectorised rows of matrix $\ma{W}_{12}$ and $\otimes$ denotes the Kronecker product.
For inclusion of the double exponential prior, it has to be rewritten as
\begin{equation}
\begin{aligned}
	\prod_{w_{ij} \in \ma{W}_{12}} \frac{1}{\sqrt{2 \pi t_{ij}}} \exp \left( -\frac{w_{ij}^2}{2 t_{ij}} \right) &= \prod_{i=1}^p \exp \left( -\frac{1}{2} (\ma{W}_{12})_{i, \cdot}^\T \ma{D}_i (\ma{W}_{12})_{i, \cdot} \right) \\
	&= \exp \left( -\frac{1}{2} vec(\ma{W}_{12}^\T)^\T \ma{D} vec(\ma{W}_{12}^\T) \right)
\end{aligned}
\end{equation}
where $(\ma{W}_{12})_{i, \cdot}$ denotes the $i$th row of $\ma{W}_{12}$, $\ma{D}_i = \diag \left( (T_{i \cdot})^{-1} \right)$, and $\ma{D} = \diag \left( \ma{D}_1, \ldots, \ma{D}_p \right)$.
The result follows from multiplying the prior by the expanded density in Eq. (\ref{eq:covtensor}).

\section{Matrix Generalised Inverse Gaussian ($\mathcal{MGIG}$) Distribution}

$\ma{X} \in \mathbb{R}^{p \times p}$ follows a Matrix Generalised Inverse Gaussian (MGIG) distribution, if $\lambda > \frac{p-1}{2}$ and
\begin{equation}
    \ma{X} \sim \mathcal{MGIG}_{p \times p}(\lambda, \ma{A}, \ma{B}) \propto \det(\ma{X})^{-\lambda-1} \exp \trace \left( -\frac{1}{2} ( \ma{A} \ma{X} + \ma{B} \ma{X}^{-1} ) \right) .
\label{eq:post:mgigdist}
\end{equation}

\paragraph{Lemma 3}

Let $p(\ma{X}) \propto \det(\ma{X})^{-\lambda-1} \exp \trace \left( -\frac{1}{2} ( \ma{A} \ma{X} + \ma{B} \ma{X}^{-1} ) \right)$ be MGIG distributed, then $\ma{W} = \ma{X}^{-1}$ is also MGIG distributed:
\begin{equation}
    p(\ma{W}) \propto \det(\ma{W})^{\lambda-p} \exp \trace \left( -\frac{1}{2} ( \ma{A} \ma{W}^{-1} + \ma{B} \ma{W} ) \right)
\label{eq:inversemgigdist}
\end{equation}

\paragraph{Proof}

Transforming $\ma{W} = \ma{X}^{-1}$ with $\dif \ma{X} = - \ma{W}^{-1} \dif \ma{W} \ma{W}$
\begin{equation}
\begin{aligned}
    p(\ma{W}) &\propto \det(\ma{W})^{-(p+1)} \det(\ma{W}^{-1})^{-\lambda-1} \exp \trace \left( -\frac{1}{2} ( \ma{A} \ma{W}^{-1} + \ma{B} \ma{W} ) \right) \\
    &= \det(\ma{W})^{\lambda - p} \exp \trace \left( -\frac{1}{2} ( \ma{A} \ma{W}^{-1} + \ma{B} \ma{W} ) \right) .
\end{aligned}
\end{equation}

\paragraph{Theorem 1, Part 2} If $\ma{S} \sim \mathcal{W}_{p+q} \left( n, \ma{\Sigma} \right)$ is Wishart distributed, then
\begin{equation}
    \ma{W}_{11} | \ma{W}_{12}, \ma{S} \sim \mathcal{MGIG}_{p \times p} \left( \frac{n}{2} + p, \ma{W}_{12} (\ma{S}_{22} + \mai) \ma{W}_{21}, \ma{S}_{11} + \mai \right)
\label{eq:mgiginwishart}
\end{equation}
is MGIG distributed.

\paragraph{Proof} The proof is similar to \cite{butler1998generalized}. Factorising the Wishart density according to
\begin{equation}
\begin{aligned}
    & \exp \trace \left( -\frac{1}{2} \ma{W} \ma{S} \right) = \exp \trace \left( -\frac{1}{2} \left( \ma{W}_{11} \ma{S}_{11} + \ma{W}_{12} \ma{S}_{21} + \ma{W}_{21} \ma{S}_{12} + \ma{W}_{22} \ma{S}_{22} \right) \right) \\
    &= \exp \trace \left( -\frac{1}{2} \left( \ma{W}_{11} \ma{S}_{11} + \ma{W}_{12} \ma{S}_{21} + \ma{W}_{21} \ma{S}_{12} + (\ma{W}_{22} - \ma{W}_{21} \ma{W}_{11}^{-1} \ma{W}_{12} + \ma{W}_{21} \ma{W}_{11}^{-1} \ma{W}_{12}) \ma{S}_{22} \right) \right) \\
    &= \exp \trace \left( -\frac{1}{2} \left( \ma{W}_{11} \ma{S}_{11} + \ma{W}_{12} \ma{S}_{21} + \ma{W}_{21} \ma{S}_{12} + (\ma{W}_{22\cdot1}  + \ma{W}_{21} \ma{W}_{11}^{-1} \ma{W}_{12}) \ma{S}_{22} \right) \right) \\
    &= \exp \trace \left( -\frac{1}{2} \left( \ma{W}_{11} \underbrace{ \ma{S}_{11} }_{=\ma{B} } + \ma{W}_{12} \ma{S}_{21} + \ma{W}_{21} \ma{S}_{12} + \underbrace{ \ma{W}_{12} \ma{S}_{22} \ma{W}_{21} }_{=\ma{A}} \ma{W}_{11}^{-1} + \ma{W}_{22\cdot1} \ma{S}_{22} \right) \right)
\end{aligned}
\label{eq:mgigfactorisation}
\end{equation}
and
\begin{equation}
\det(\ma{W})^{\frac{n}{2}} = \det(\ma{W}_{11})^{\frac{n}{2}} \det(\ma{W}_{22\cdot1})^{\frac{n}{2}}
\label{eq:mgigdeterminant}
\end{equation}
where we changed variables $\ma{W}_{22\cdot1} = \ma{W}_{22} - \ma{W}_{21} \ma{W}_{11}^{-1} \ma{W}_{12}$ with Jacobian $1$ and integrated over $\ma{W}_{22\cdot1}$.
Comparing Eq. (\ref{eq:inversemgigdist}) with Eq. (\ref{eq:mgigfactorisation}), we can identify $\ma{A}$, $\ma{B}$, and $\lambda$:
\begin{equation}
    \frac{n}{2} \stackrel{!}{=} \lambda - p \quad \Rightarrow \quad \lambda = \frac{n}{2} + p.
\end{equation}
such that
\begin{equation}
    \ma{W}_{11} | \ma{W}_{12}, \ma{S} \sim \mathcal{MGIG}_{p \times p} \left( \frac{n}{2} + p, \ma{W}_{12} \ma{S}_{22} \ma{W}_{21}, \ma{S}_{11} \right)
\end{equation}
Since the double exponential prior does not affect the distribution of $\ma{W}_{11}$, we only have to include the Wishart prior. The result follows immediately.

\newpage
\small{
    \bibliography{arxiv2015_markov_blanket}
    \bibliographystyle{plain}
}

\end{document}